\documentclass[conference]{IEEEtran}
\IEEEoverridecommandlockouts
\usepackage{cite}
\usepackage{amsmath,amssymb,amsfonts}
\usepackage{algorithmic}
\usepackage{graphicx}
\usepackage{textcomp}
\usepackage{xcolor}
\usepackage[hyphens]{url}

\def\eg{\emph{e.g}}
\def\etal{\emph{et al}}

\usepackage{hyperref}
\hypersetup{colorlinks,linkcolor={red},citecolor={red},urlcolor={red}}

\def\BibTeX{{\rm B\kern-.05em{\sc i\kern-.025em b}\kern-.08em
    T\kern-.1667em\lower.7ex\hbox{E}\kern-.125emX}}
\begin{document}

\title{In Ictu Oculi: Exposing AI Generated Fake Face Videos by Detecting Eye Blinking
}
\author{Yuezun Li, Ming-Ching Chang and Siwei Lyu\footnote{Corresponding Author} \\
Computer Science Department, University at Albany, SUNY}
\maketitle

\begin{abstract}
The new developments in deep generative networks have significantly improve the quality and efficiency in generating realistically-looking fake face videos. In this work, we describe a new method to expose fake face videos generated with neural networks. Our method is based on detection of eye blinking in the videos, which is a physiological signal that is not well presented in the synthesized fake videos. Our method is tested over benchmarks of eye-blinking detection datasets and also show promising performance on detecting videos generated with DeepFake.

\end{abstract}


\section{Introduction}

The advancement in camera technology, wide availability of cellphones and increasing popularity of social networks (FaceBook, Twitter, WhatsApp, InstaGram, and SnapChat) and video sharing portals (YouTube and Vemeo) have made the creation, editing and propagation of digital videos more convenient than ever. This has also bring forth digital tampering of videos as an effective way to propagate falsified information. 

Unlike the case for digital images, editing videos has been a time-consuming and painstaking task due to the lack of sophisticated editing tools like PhotoShop and the large number of editing operations involved for a video -- as a case in point, a 20 second video with 25 frames per second requires editing of 500 images. As such, highly realistic fake videos were rare, and most can be identified relatively easily based on some conspicuous visual artifacts. 

However, the situation has been changed dramatically with the new generation of generative deep neural networks \cite{liu2017unsupervised}, which are capable of synthesizing videos from large volume of training data with minimum manual editing. The situation first caught attention in earlier 2018, when a software tool known as {\tt DeepFake} was made publicly available based on such an approach. DeepFake uses generative adversary networks (GANs) to create videos of human faces of one subject to replace those in an original video. Because the GAN models were trained using tens of thousands of images, it can generate realistic faces that can be seamlessly spliced into the original video, and the generated video can lead to falsification of the subject's identity in the video. Subsequently, there had been a surge of fake videos generated using this tool and uploaded to YouTube for gross violations of privacy and identity, some with serious legal implications \footnote{For example, see \url{https://www.lawfareblog.com/deep-fakes-looming-crisis-national-security-democracy-and-privacy}.}. 

Detecting such fake videos becomes a pressing need for the research community of digital media forensics. 

While traditional media forensic methods based on cues at the signal level (\eg, sensor noise, CFA interpolation and double JPEG compression), physical level (\eg, lighting condition, shadow and reflection) or semantic level (\eg, consistency of meta-data) can be applied for this purpose, the AI generated fake face videos pose challenges to these techniques. 
In this work, we describe a first forensic method targeting at such fake videos. The general methodology we follow is to detect the lack of physiological signals intrinsic to human beings that are not well captured in the synthesized videos. Such signals may include spontaneous and involuntary physiological activities such as breathing, pulse and eye movement, and are oftentimes overlooked in the synthesis process of fake videos. Here, we focus on the detection of the lack of {\em eye blinking} to expose AI synthesized face videos. Our method is based on a novel deep learning model combining a convolutional neural network (CNN) with a recursive neural network (RNN) to capture the phenomenological and temporal regularities in the process of eye blinking. Current methods employ Convolutional Neural Networks (CNN) as a binary classifier to distinguish open and close eye state of each frame. However, CNN generates predictions based on single frame, which does not leverage the knowledge in temporal domain. As human eye blinking has strongly temporal correlation with previous states, we employ Long-term Recurrent Convolutional Neural Networks (LRCN) \cite{donahue2015long} distinguish open and close eye state with the consideration of previous temporal knowledge. Our method is tested over benchmarks of eye-blinking detection datasets and also show promising performance on detecting videos generated with DeepFake.
\begin{figure*}[t]
 \centering
 \includegraphics[width=\linewidth]{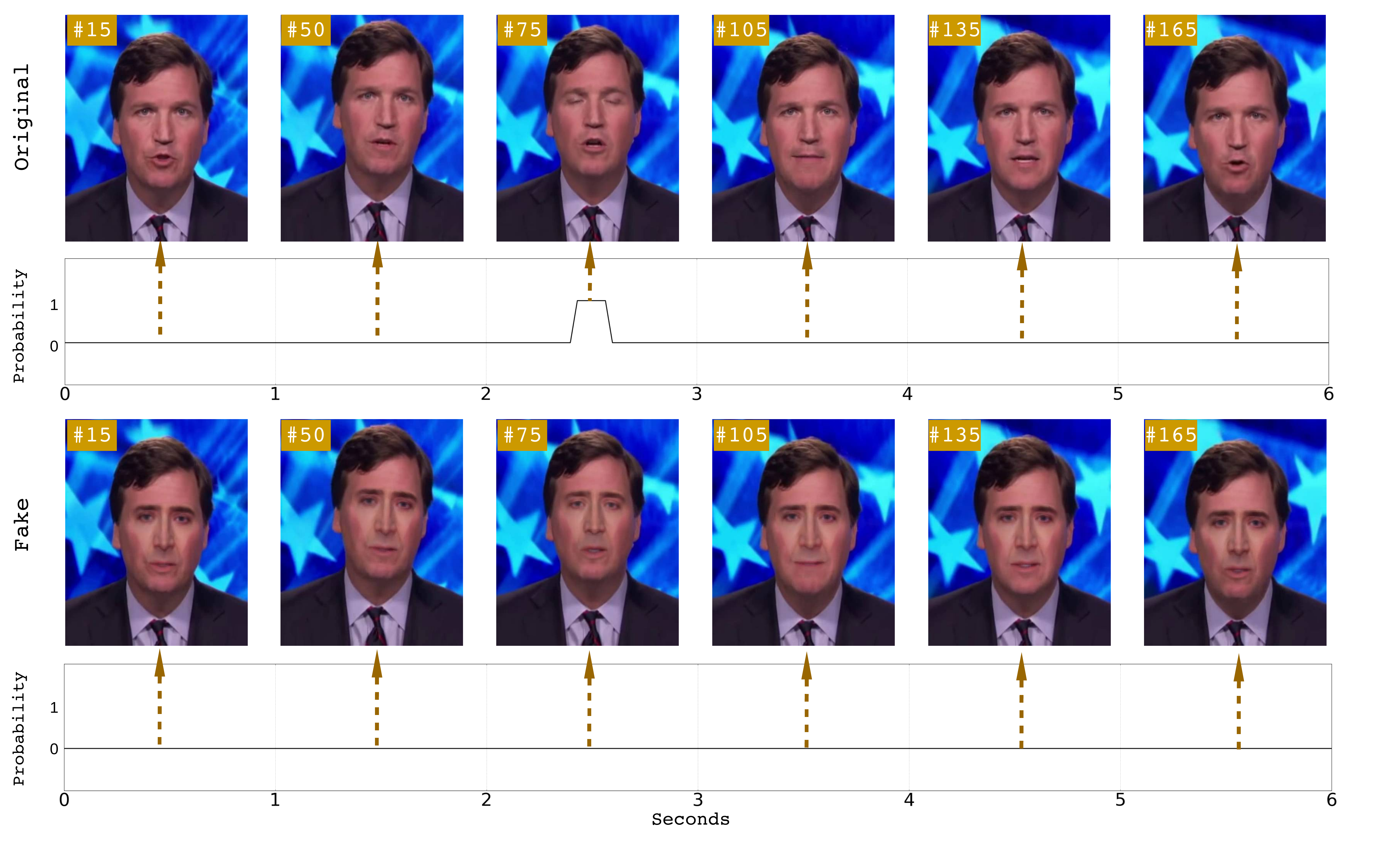}
 ~\vspace{-3em}
  \caption{\small \em Example of eye blinking detection on an original video ({\bf top}) and a DeepFake generated fake video ({\bf bottom}. Note that in the former, an eye blinking can be detected within 6 seconds, while in the latter such is not the case, which is abnormal from the physiological point of view.}
  ~\vspace{-2em}
\end{figure*}

\section{Backgrounds and Related Works}

Blinking refers to the rapid closing and opening movement of the eyelid. There are three major types of blinking,  as Spontaneous blink, Reflex blink and Voluntary blink. The Spontaneous blink refers to blinking without external stimuli and internal effort, which is controlled  the pre-motor brain stem and happens without conscious effort, like breathing and digestion. Spontaneous blinking serves an important biological function that moisturizes with tears and remove irritants from the surface of the cornea and conjunctiva.  For a health adult human, generally, between each blink is an interval of 2–10 seconds but the actual rates vary by individual. 

The mean resting blinking rate is 17 blinks/min or 0.283 blinks per second\footnote{\url{https://www.ncbi.nlm.nih.gov/pubmed/9399231.}}(during conversation this rate increases to 26 blinks/min, and decreases to 4.5 blinks/second while reading — this difference may be interesting in our analysis since many of the talking-head politicians are probably reading when they are being filmed).  For now, however, I’ll assume an average rate of 17 blinks/min. The length of a blink is 0.1-0.4 seconds/blink\footnote{\url{http://bionumbers.hms.harvard.edu/bionumber.aspx?id=100706\&ver=0}.}. 

AI generated face lack eye blinking function, as most training datasets do not contain faces with eyes closed. The lack of eye blinking is thus a telltale sign of a video coming from a different source than a video recorder. If we assume an average exposure time of 1/30 second then I estimate that the probability of capturing a photo with someone blinking is about 7.5\%. Of course, most photos of someone online won’t show them with their eyes closed, so this likelihood goes way down in practice.

\subsection{AI Generation of Fake Videos}

Realistic images/videos have been generated using detailed 3D computer graphics models. recently, the development of new deep learning algorithms, especially those based on the generative adversary networks (GANs). Goodfellow \etal. \cite{goodfellow2014generative} first proposed generative adversarial networks (GANs), which typically consist of two networks: generator and discriminator. The generator aims to produce a sample which should not be distinguished from training data distribution, while discriminator is to assess the sample produced by generator. Denton \etal \cite{denton2015deep} proposed a Laplacian pyramid GAN to generate images in
a coarse-to-fine fashion. Radford \etal. \cite{radford2015unsupervised} proposed Deep Convolutional GANs (DCGAN) and showed the potential for unsupervised learning. Arjovsky \etal. \cite{arjovsky2017wasserstein} used Wasserstein distance to make training stable. Isola \etal. \cite{isola2017image} investigated conditional adversarial networks to learn mapping from input image to output image and also the loss function to train the mapping. Taigman \etal. \cite{taigman2016unsupervised} proposed the Domain Transfer Network (DTN) to map a sample from one domain to an analog sample in another domain and achieved favorable performance on small resolution face and digit images. Shrivastava \etal. \cite{shrivastava2017learning} reduced the gap between synthetic and real image distribution using a combination of adversarial loss and self-regularization loss. Liu \etal. \cite{liu2017unsupervised} proposed an unsupervised image to image translation framework based on coupled GANs, with the aim to learn the joint distribution of images in different domains.

\subsection{Eye Blinking Detection}

Detecting eye blinking has been studied in computer visions for applications in fatigue detection \cite{horng2004driver,wang2006driver,dong2005fatigue,azim2014fully,mandal2017towards} and face spoof detection \cite{boulkenafet2015face,galbally2014face,li2017face,steiner2016reliable,li2018learning}, and various methods have been proposed to solve this problem. 

Pan \etal. \cite{pan2007eyeblink} constructs undirected conditional random field
framework to infer eye closeness such that eye blinking is detected. 
Sukno \etal. \cite{sukno2009automatic} employs Active Shape Models with the Invariant Optimal Features to delineate the outline of eyes and computes the eye vertical distance to decide eye state.  
Torricelli \etal. \cite{torricelli2009adaptive} utilizes the difference between consecutive frames to analyze state of eyes. 
Divjak \etal. \cite{divjak2009eye} employs optical flow to obtain eye movement and extract the dominant vertical eye movement for blinking analysis.
Yang \etal. \cite{yang2012robust} models the shape of eyes
based on a pair of parameterized parabolic curves, and fit the
model in each frame to track eyelid.
Drutarovsky \etal. \cite{drutarovsky2014eye} analyzes the variance of the vertical motions of eye region which is detected by a Viola–Jones type algorithm. Then a
flock of KLT trackers is used on the eye region. Each eye region is divided into 3x3 cells and an average motion in each cell is calculated.
Soukupova \etal. \cite{soukupova2016real} fully relies on face landmarks and proposes a single scalar quantity -- eye aspect ratio (EAR), to characterize the eye state in each frame. Then a SVM is trained using EAR values within a short time window to classify final eye state. 
Kim \etal \cite{kim2017study} studies CNN-based classifiers to detect eye open and close state. They adopt ResNet-50 \cite{he2016deep} model and compare the performance with AlexNet and GoogleNet. 

In this paper, we extend the work on CNN-based classifier to LRCN \cite{donahue2015long}, which incorporates the temporal relationship between consecutive frames, as eye blinking is a temporal process which is from opening to closed, such that LRCN memorize the long term dynamics to remedy the effect by noise introduced from single image.  

\begin{figure*}[t]
 \centering
 \includegraphics[width=\linewidth]{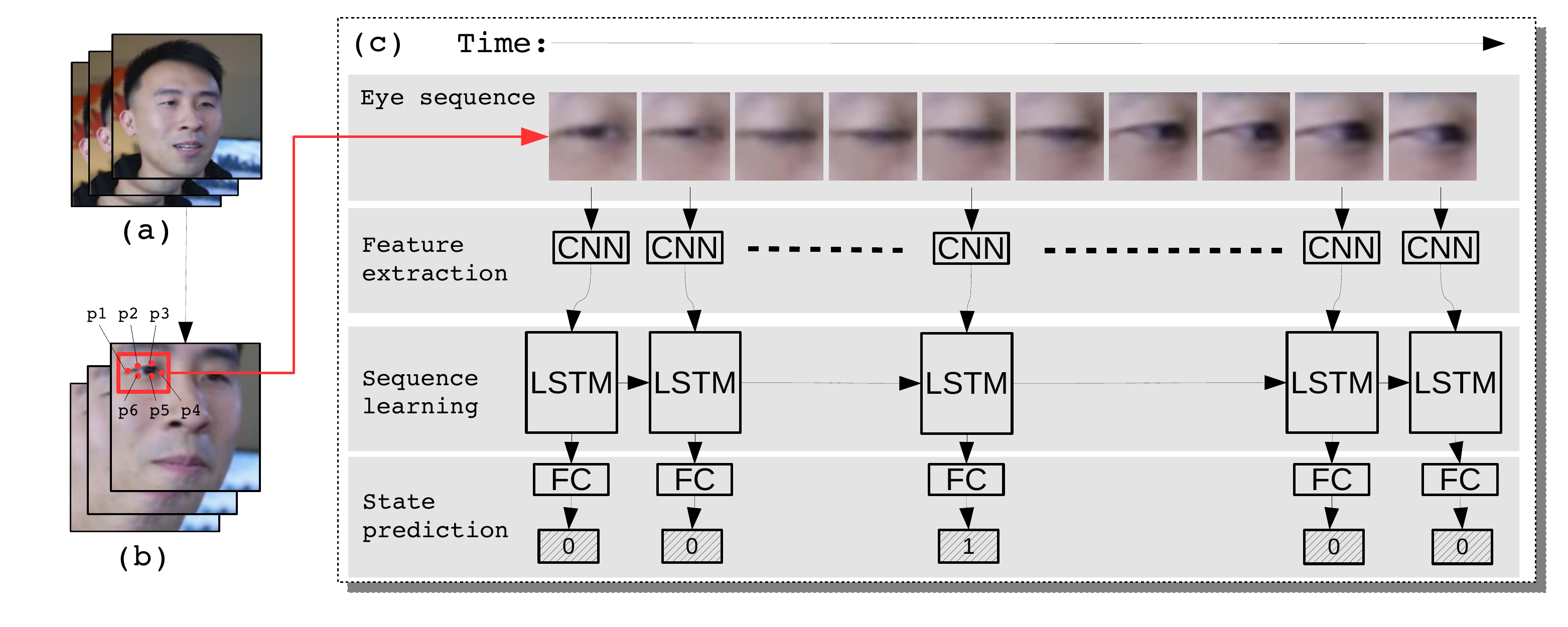}
~\vspace{-3em}
\caption{\em \small Overview of our LRCN method. (a) is the original sequence. (b) is the sequence after face alignment. We crop out eye region of each frame based on eye landmarks $p_{1 \sim 6}$ in (b) and pass it to (c) LRCN, which consists of three parts: feature extraction, sequence learning and state prediction.}
~\vspace{-2em}
   \label{fig:overview}
\end{figure*}

\section{Method}
\label{sec:method}

In this section, we describe in detail our method to detecting eye blinking in a video. 
The general pipeline of our algorithm is provided in Figure \ref{fig:overview}. Our method first detects faces in each frame of the video\footnote{For simplicity, we focus on one subject shown in each frame of the video to analyze, but this can be easily extended to the case of multiple subjects in the video.} The detected faces are then aligned into the same coordinate system to discount head movements and changes in orientations based on detection of facial landmark points. Regions corresponding to each eye are extracted out to form a stable sequence (top row in Fig.\ref{fig:overview}(c)). After these pre-processing steps, eye blinking is detected by quantifying the degree of openness of an eye in each frame in video using the LRCN model. We describe each of these steps in sequel.

\subsection{Pre-processing}

The first step in our method is to locate the face areas in each frame of the video using a face detector. Then facial landmarks, which are locations on the face carrying important structural information such as tip of the eyes, noses and mouths and contours of the cheek, are extracted from each detected face area. 

The head movement and changes in face orientation in the video frames introduce distractions in facial analysis. As such, we first align the face regions to a unified coordinate space using landmark based face alignment algorithms. Specifically, given a set of face landmarks in original coordinate, 2D face alignment is to warp and transform the image to another coordinate space, where the transformed face is (1) in the center of image, (2) rotated to make eyes lie on a horizontal line and (3) scaled to a similar size. 

From the aligned face areas, we can extract a surrounding rectangular regions of the landmarks corresponding to the contours of the eyes into a new sequence of input frames, see Figure \ref{fig:overview}(b). Specifically, the rectangle region is generated by first extracting the bounding boxes of each eye's landmark points, then enlarging the bounding box by $(1.25,1.75)$ in the horizontal and vertical directions. This is to guarantee that the eye region is included in the cropped region. The cropped eye area sequences are passed into LRCN for dynamic state prediction.

\subsection{Long-term Recurrent Convolutional Networks (LRCN)}
\label{sec:lrcn}

As human eye blinking shows strong temporal dependencies, we employ the Long-term Recurrent Convolutional Networks (LRCN) model \cite{donahue2015long} to capture such temporal dependencies.

\begin{figure}[t]
 \centering
 \includegraphics[width=0.8\linewidth]{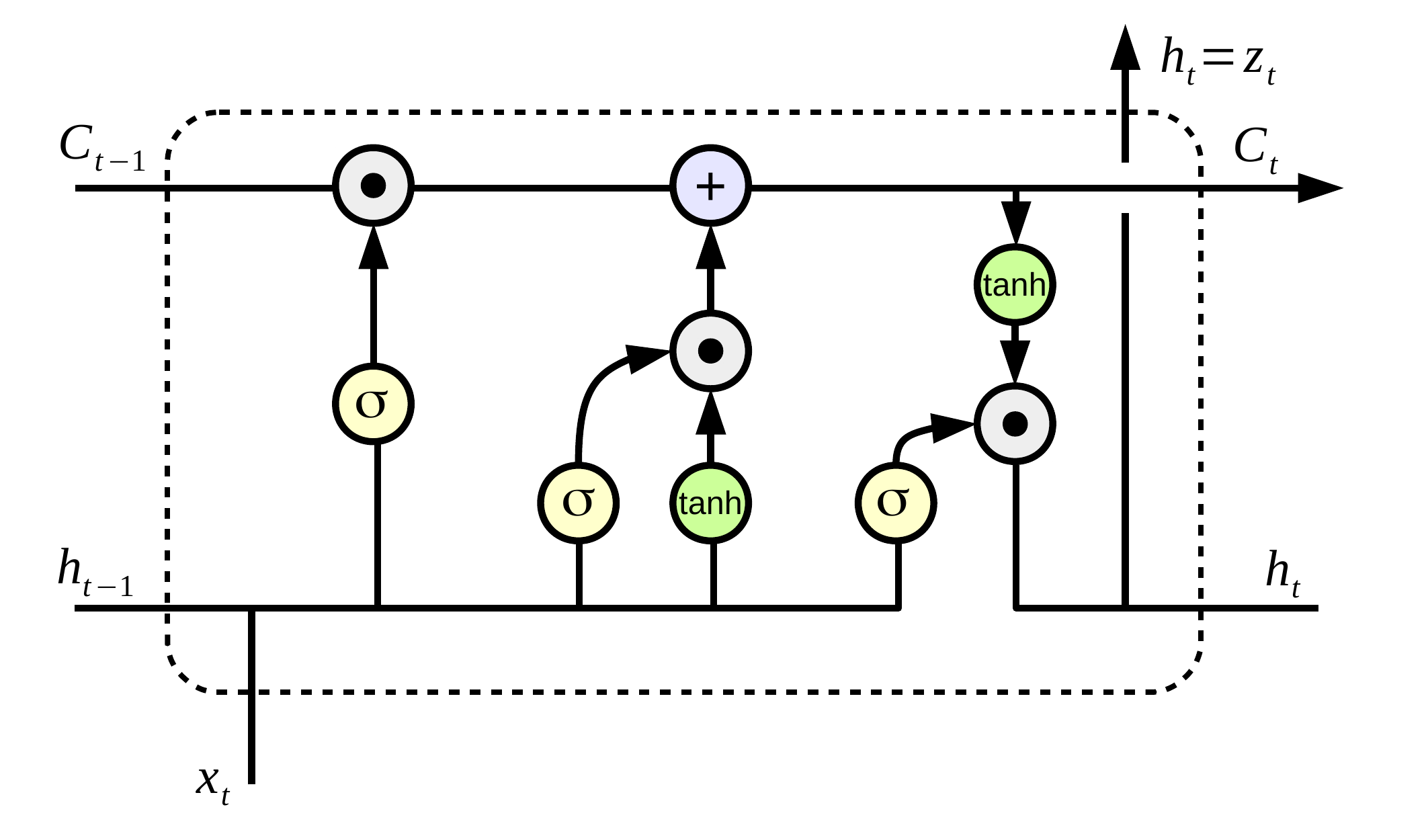}
  \caption{\em A diagram of LSTM structure.}
   \label{fig:lstm}
\end{figure}

As shown in Figure \ref{fig:overview}(c), the LRCN model is composed by three parts, namely, {\em feature extraction}, {\em sequence learning} and {\em state prediction}. 
Feature extraction module convert the input eye region into discriminative features. It is implemented with a Convolutional Neural Network (CNN) based on the VGG16 framework \cite{simonyan2014very} but without \texttt{fc7} and \texttt{fc8} layers\footnote{Other deep CNN architecture such as ResNet \cite{he2016deep} can also be used but for simplicity we choose VGG16 in the current work.}. VGG16 is composed by five blocks of consecutive convolutional layers \texttt{conv1} $\sim$ \texttt{5}, where max-pooling operation follows each block. Then three fully connected layers \texttt{fc6} $\sim$ \texttt{8} are appended on the last block.
The output from the feature extraction is fed into sequence learning, which is implemented with a recursive neural network (RNN) with Long Short Term Memory (LSTM) cells \cite{Hochreiter:1997}. The use of LSTM-RNN is to increase the memory capacity of the RNN model and avoid the gradient vanishing in the back-propagation-through-time (BPTT) algorithm in the training phase.  LSTMs are memory units that control when and how to forget previous hidden states and when and how to update hidden states. We use LSTM as illustrated in Figure \ref{fig:lstm}, where $\sigma(x) = \frac{1}{1+e^{-x}}$ is {\em sigmoid} function to push input into $[0,1]$ range, $tanh(x) = \frac{e^x - e^{-x}}{e^x+e^{-x}}$ is {\em hyperbolic tangent} function which squash input into $[-1, 1]$ range, $\odot$ denotes inner product of two vectors. Given input $C_{t-1}, h_{t-1}, x_t$, the LSMT updates along with time $t$ by

\begin{equation}
\begin{array}{ll}
f_t = \sigma(W_{fh} h_{t-1} + W_{fx}  x_t + b_f) \\
i_t = \sigma(W_{ih}  h_{t-1} + W_{ix}  x_t + b_i) \\
g_t = tanh(W_{ch}  h_{t-1} + W_{cx}  x_t + b_c) \\
C_t = f_t 	\odot  C_{t-1} + i_t \odot  g_t \\
o_t = \sigma(W_{oh}h_{t-1} + W_{ox}x_t +  b_o) \\
h_t = o_t \odot tanh(C_t)
\end{array}
\end{equation}
where $f_t$ is forget gate to control what previous memories will be discard, $i_t$ is input gate to selectively pass the current input, which is manipulated by $g_t$, $o_t$ is output gate to control how much memory will be transferred into hidden state $h_t$. Memory cell $C_t$ is combined by previous memory cell $C_{t-1}$ controlled by $f_t$ and manipulated input $g_t$ controlled by $i_t$. We use $256$ hidden units in LSTM cell, which is the dimension of LSTM output $z_t$. 

For the final state prediction stage, the output of each RNN neuron is further sent to neural network consists of a fully connected layer, which takes the output of LSTM and generate the probability of eye open and close state, denoted by $0$ and $1$ respectively.

\subsection{Model Training}
\label{subsec:training}

The training of the LRCN model is performed in two steps. In the first step, we train the VGG based CNN model based on a set of labeled training data consisting of eye regions corresponding to open and closed eyes. The model is trained using back-propagation implemented with stochastic gradient descent {and dropout by probability $0.5$ in fully connected layers}. In the second step, the LSTM-RNN and fully connected part of the network are trained jointly using the back-propagation-through-time (BPTT) algorithm. In both cases, the loss objective is cross entropy loss with binary classes (open or closed).


\section{Experiments}
We train the LRCN model based on image datasets of eye open states. We then test the algorithm detecting eye blinking on authentic and fake videos generated with the DeepFake algorithm.

\subsection{Datasets}
\label{subsec:dataset}

To date, there are a few image datasets that can be used for evaluating algorithms that detect closed eyes, such as the CEW Dataset \cite{song2014eyes}\footnote{\url{http://parnec.nuaa.edu.cn/xtan/data/ClosedEyeDatabases.html.}}, which includes $1,193$ images of closed eyes and $1,232$ images of open eyes\footnote{The other dataset, the EEG Eye State Data Set \url{https://archive.ics.uci.edu/ml/datasets/EEG+Eye+State}, is not available to download.} However, no existing video dataset specially designed for the same 
purpose is available\footnote{There is the ZJU Eyeblink Video Database \cite{pan2007eyeblink}, but there is no access to the data.}, which is important due to the temporal nature of eye blinking. To be able to experimentally evaluate our algorithm, we downloaded $50$ videos, where each represents one individual and lasts approximate $30$ seconds with at least one blinking occurred, to form the Eye Blinking Video (EBV) dataset. 
We annotate the left and right eye states of each frame of the videos using a user-friendly annotation tool we developed. Our dataset is available to download from \url{http://www.cs.albany.edu/~lsw/downloads.html}. In our experiments, we use the CEW dataset to train the front-end CNN model, and select 40 videos as our training set for the overall LRCN model and 10 videos as the testing set.

\subsection{Generating Fake Videos}
\begin{figure*}[t]
 \centering
 \includegraphics[width=\linewidth]{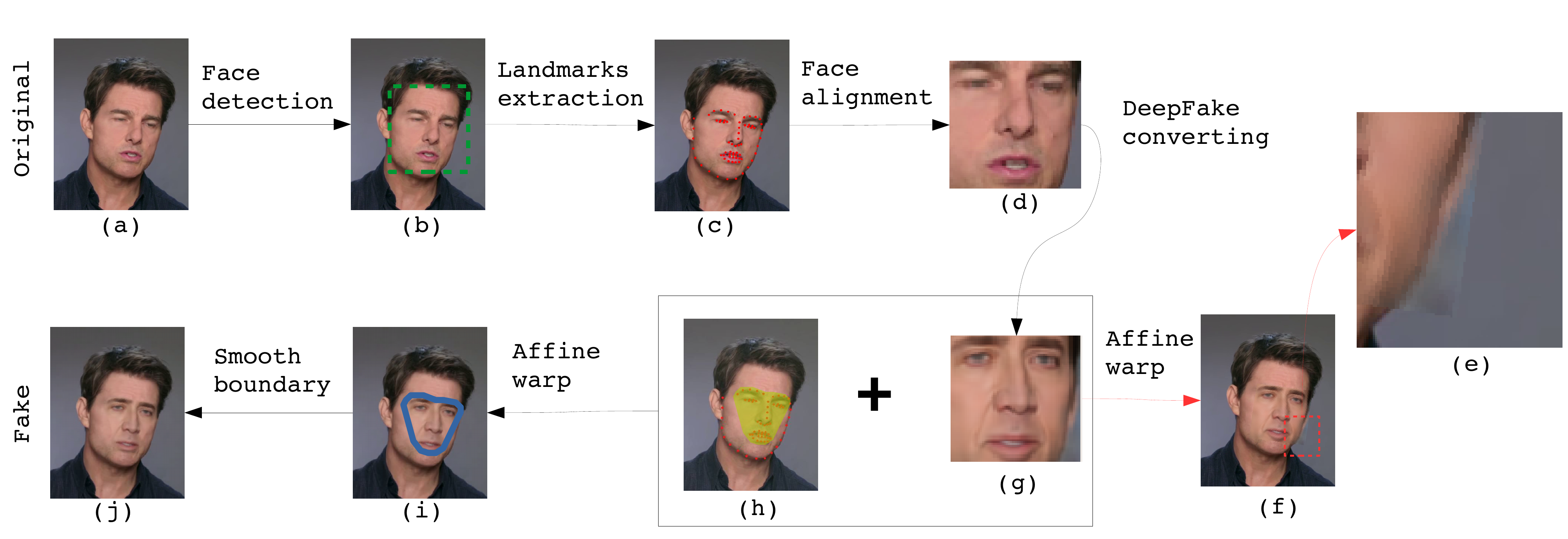}
  \caption{\em Overview of fake face generation pipeline. (a) is the original input image. The green dash box in (b) is face area localized by face detector. (c) is the face landmarks. (d) is the face after alignment. DeepFake takes (d) as input and convert it to (g). The artifacts are introduced by directly affine warping generated face back to (a), as shown in (f, e). (h) denotes the convex polygon mask that generated face inside is retained. (i) is the smooth process on boundary of mask. (i) is the final fake image.}
   \label{fig:fake_face}
\end{figure*}

We use DeepFake with post-processing to generate fake face videos, see Figure \ref{fig:fake_face}. Specifically, we first use \texttt{dlib} to detect face area in each image. Then face landmarks are extracted for face alignment as described in section \ref{sec:method}. We then generate the corresponding fake faces using the DeepFake algorithm. If we directly affine warp this rectangle of fake face back to image using similarity transformation matrix, the boundary of rectangle is visible in most cases as the slight color difference of real and fake face area, as shown in Figure \ref{fig:fake_face}(e). To reduce such artifacts, we generate a specific mask which is a convex polygon determined by landmarks of left and right eyebrow, and the bottom mouth. As such, we only retain content inside this mask after affine warping fake face back to original image. To further smooth the transformation, we apply Gaussian blur to the boundary of mask. With this procdure, we generate $49$ fake videos.

\subsubsection{Data preparation} 

Face detection, landmark extraction and face alignment are implemented based on  library \texttt{dlib} \cite{dlib09}, which integrates the current state-of-the-art face analysis algorithms.  We generate eye sequences by cropping out eye area of each frame of our video dataset. 

We augment data to increase training robustness. The training of the front-end CNN model takes images as input, so we use each frame of generated eye sequences as training sample, with additional augmentation: horizontal flipping image, modifying image color contrast, brightness and color distortion. For LRCN joint training, eye sequences are required. In particular, the augmentation process for sequence should be consistent to avoid affect temporal relationship, such that the process for each frame in sequence should be same.

With combination of our cropped eye images and CEW dataset, we train VGG16 as a binary image classifier to distinguish eye state in image domain. The input size is fixed as $224$x$224$ and the batch size is $16$. The learning rate starts from $0.01$ and decays by $0.9$ each $2$ epochs. We employ stochastic gradient descent optimizer and terminate training until it reaches the maximum epoch number $100$. Then we remove \texttt{fc7}, \texttt{fc8} layers from trained VGG16 to be the feature extraction part of LRCN. 

We randomly select a sequence which contains a variety of temporal consecutive eye images with at least one blinking occurred as LRCN input. Each sample has variable length between $10$ to $20$ images.
We fix the parameters of CNN layers we obtain above and perform training on rest part: LSTM cells and \texttt{fc} layer. We set batch size as $4$. The learning rate starts from $0.01$ and decay by $0.9$ each $2$ epochs. We use the ADAM optimizer \cite{kingma2014adam} and terminate training until $100$ epochs.  

\subsection{Evaluations}
\label{subsec:evaluations}
\begin{figure}[t]
 \centering
 \includegraphics[width=0.8\linewidth]{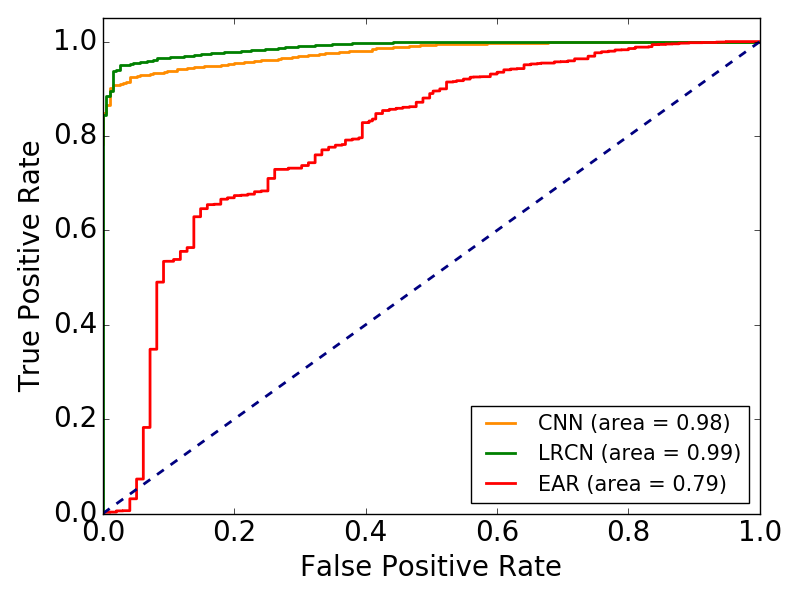}
  \caption{\em Illustration of ROC curve for CNN, LRCN and EAR.}
   \label{fig:roc}
\end{figure}

We evaluate our LRCN method with comparison to other methods: Eye Aspect Ratio (EAR) \cite{soukupova2016real} and CNN. CNN image classifier is trained on image domain to distinguish different classes. We employ VGG16 as our CNN model to distinguish eye state. 

EAR method replies on eye landmarks to analyze eye state, in terms of the ratio between the distance of upper and lower lid, and the distance between left and right corner point, which is defined as $\frac{\|p_2 - p_6 \| + \|p_3 - p_5 \|}{2\|p_1 - p_4 \|}$ (see Figure \ref{fig:overview}(b)). This method runs fast as merely cost in ratio computation. However, the main drawback of EAR method is that it fully depends on eye landmarks, which is not reliable in many cases. 

\begin{figure*}[t]
 \centering
 \includegraphics[width=\linewidth]{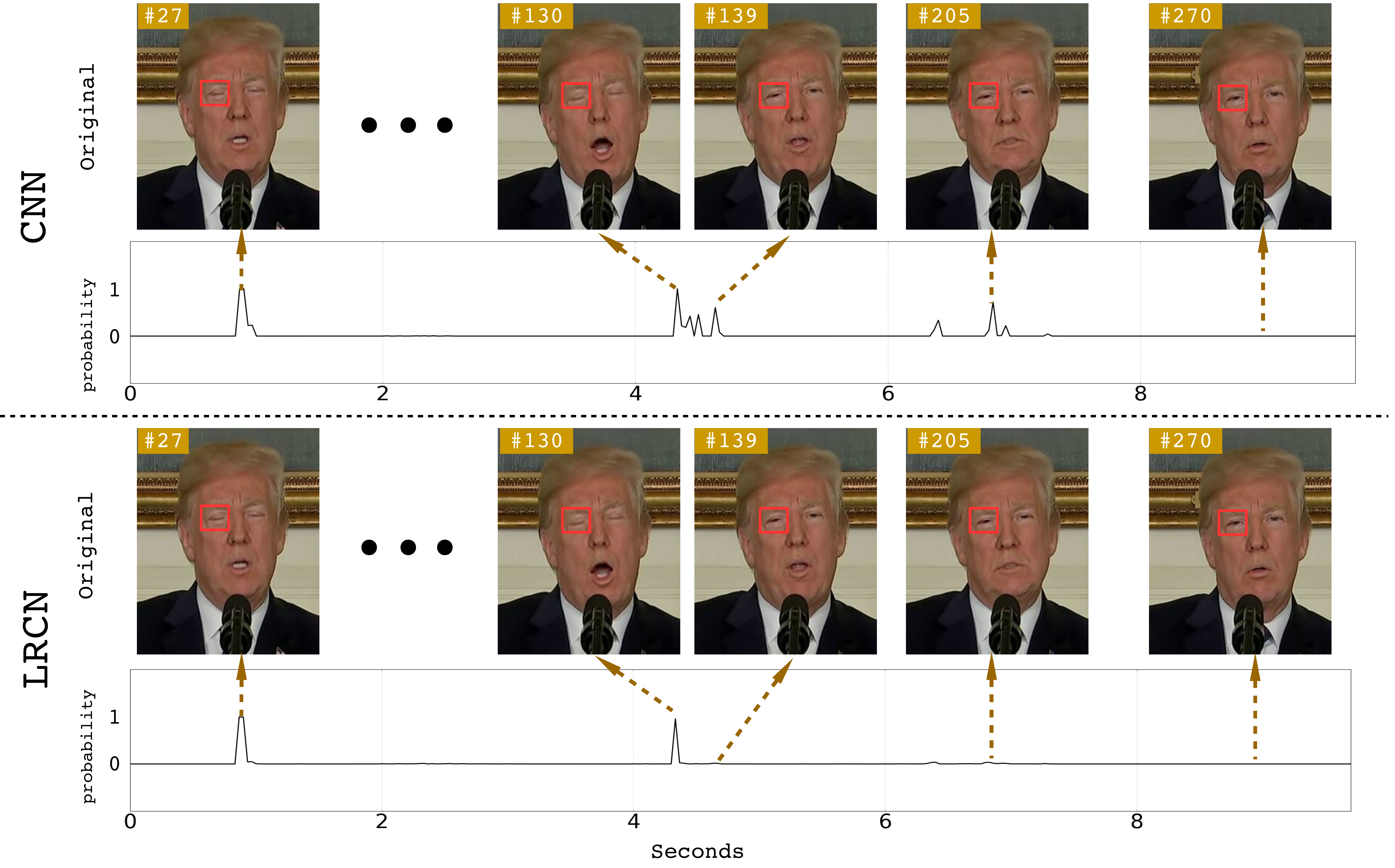}
  \caption{\em Illustration of comparing CNN and LRCN on left eye of Trump video. LRCN exhibits more smooth and accurate results than CNN}
   \label{fig:cnn_vs_lrcn}
\end{figure*}

{\flushleft \bf Main results:} We evaluate these three methods on our own testing data mentioned in section \ref{subsec:dataset}. Figure \ref{fig:roc} illustrate the ROC curve of three methods. Observe that LRCN show the best performance $0.99$ compared to CNN $0.98$ and EAR $0.79$. CNN in this experiment shows an exceptional well performance to distinguish the eye state on image domain. However, its prediction does not consider temporal knowledge of previous state. LRCN takes advantage of long term dynamics to effectively predict eye state, such that it is more smooth and accurate. An example is illustrated in Figure \ref{fig:cnn_vs_lrcn}. At some frames, the state of left eye is ambiguous since eye area is small. Only depending on image domain, CNN will be confused. With temporal domain, LRCN can memorize the previous state. If blinking has occurred before, the eye state in next couple frames are very likely to be open (Figure \ref{fig:cnn_vs_lrcn} \#139). If there is no trend of eye closing before, the eye state of next frame is very likely to be open (Figure \ref{fig:cnn_vs_lrcn} \#205).

\section{Conclusion}

The new developments in deep generative networks have significantly improve the quality and efficiency in generating realistically-looking fake face videos. In this work, we describe a new method to expose fake face videos generated with neural networks. Our method is based on detection of eye blinking in the videos, which is a physiological signal that is not well presented in the synthesized fake videos. Our method is tested over benchmarks of eye-blinking detection datasets and also show promising performance on detecting videos generated with DeepFake.

There are several directions that we would like to further improve the current work. First, we will  explore other deep neural network architectures for more effective methods to detect closed eyes. Second, our current method only uses the lack of blinking as a cue for detection. However, the dynamic pattern of blinking should also be considered -- too fast or frequent blinking that is deemed physiologically unlikely could also be a sign of tampering. Finally, eye blinking is a relatively easy cue in detecting fake face videos, and sophisticated forgers can still create realistic blinking effects with post-processing and more advanced models and more training data.  So in the long run, we are interested in exploring other types of physiological signals to detect fake videos.

\bibliographystyle{IEEEtran}
\bibliography{ref}

\end{document}